\documentclass[runningheads]{llncs}
\usepackage{graphicx}
\usepackage{amsmath,amssymb} 
\usepackage{color}
\usepackage[width=122mm,left=12mm,paperwidth=146mm,height=193mm,top=12mm,paperheight=217mm]{geometry}
\usepackage{hyperref}

\newcommand{\argmin}[1]{\underset{#1}{\operatorname{arg}\,\operatorname{min}}\;}

\begin{document}
\pagestyle{headings}
\mainmatter

\title{Shape from Mixed Polarization} 

\author{Vage Taamazyan\inst{1} \and Achuta Kadambi\inst{2} \and Ramesh Raskar\inst{2}}

\institute{Moscow Institue of Physics and Technology \and MIT Media Lab \\
\email{vaheta@tardis3d.ru} \hspace{25pt} \email{\{achoo,raskar\}@mit.edu} }

\maketitle

\begin{abstract}
Shape from Polarization (SfP) estimates surface normals using photos captured at different polarizer rotations. Fundamentally, the SfP model assumes that light is reflected either diffusely or specularly. However, this model is not valid for many real-world surfaces exhibiting a mixture of diffuse and specular properties. To address this challenge, previous methods have used a sequential solution: first, use an existing algorithm to separate the scene into diffuse and specular components, then apply the appropriate SfP model. In this paper, we propose a new method that jointly uses
viewpoint and polarization data to holistically separate
diffuse and specular components, recover refractive index, and ultimately recover 3D shape. By involving the physics of polarization in the separation process, we demonstrate
competitive results with a benchmark method, while recovering additional information (e.g. refractive index).
\end{abstract}

\section{Introduction}

For centuries, it has been known that the shape of an object influences the polarization state of reflected light.\footnote{Augustin-Jean Fresnel (1788-1827).} This principle underlies the Shape from Polarization (SfP) technique, which aims to recover the surface normals of an object from three polarized photos. 

Classical approaches to SfP rely on specular reflections from an object (hereafter, specular SfP). In an effort to handle purely diffuse surfaces, Atkinson and Hancock introduced a landmark result, modifying the physical model to account for cases where all the light is diffusely reflected (hereafter, diffuse SfP)~\cite{atkinson2006recovery}. However, many surfaces exhibit properties that are neither diffuse nor specular, but somewhere in-between. A ``mixed reflection'' occurs: both diffusely and specularly reflected light return to the camera causing model mismatch. 

Obtaining surface normals through polarization is mostly a laboratory problem, with several practical challenges. For example, one needs to know the refractive index of the material; the material must be either diffuse or specular; and ill-posed ambiguities exist for both zenith and azimuth angles. Recent work has used a coarse depth map to provide what may be a promising step toward ``in-the-wild'' uses of SfP~\cite{kadambi2015polarized} (hereafter, ``Polarized 3D''). While Polarized 3D has demonstrated compelling results, we believe our work offers complementary benefits.

At the heart of our work is an analysis of mixed reflections and their impact on existing techniques that use SfP. We find that, indeed, a mixed reflection perturbs the result to the point where correction is desirable. We therefore propose a physics-based technique to correct for mixed reflections using multiple viewpoints of an object, demonstrating the practical benefits of our approach through comparisons with previous work. 

\paragraph{Scope:} Our contribution of extending SfP to handle mixed surfaces is a unified approach. Prior art has proposed a sequential approach, where the scene is first split into diffuse and specular components, following which the appropriate SfP algorithm can be used. For example, the work of Miyazaki et al.~\cite{miyazaki2003polarization} handles mixed surfaces by first using an algorithm for diffuse-specular separation, proposed by Tan and Ikeuchi~\cite{tan2005separating}, following which standard technique of SfP are used. Since the Tan and Ikeuchi technique is very general, i.e, it is not specific to polarization, we believe that the information from the Fresnel equations could be used to improve on previous work. 

In this paper, we develop an approach that incorporates the SfP model to aid in separating the image into diffuse and specular components. We also show that our proposed approach allows simultaneous recovery of refractive index, while outperforming sequential approaches. 

\begin{figure}[t]%
\centering
\includegraphics[width=1\columnwidth]{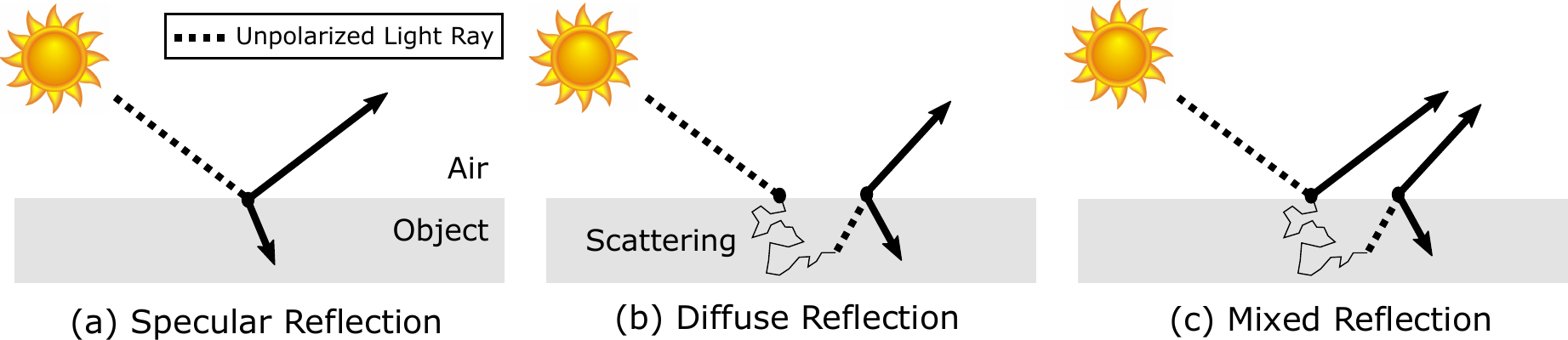}%
\caption{(a) Unpolarized light strikes a surface -- the specular reflection is partially polarized in accordance with the Fresnel equations. (b) The Fresnel equations were recently extended to the case of diffuse reflections by Atkinson and Hancock~\cite{atkinson2006recovery}, where it was assumed that the light emanating from the object is transmitted from an internal reflection. (c) Mixed reflections are the subject of this paper.}%
\label{fig:refl}%
\end{figure}



\section{Related Work}

In the context of related work, we believe our proposed technique is the first unified approach toward joint estimation of shape, diffuse-specular separation, and pixel-wise refractive index.

\paragraph{The Fresnel Equations} describe the behavior of electromagnetic radiation as it interacts with a surface. When light interacts with a surface, the first-order event that occurs is reflection and transmission at the boundary (Figure \ref{fig:refl}a).\footnote{``First-order'' refers to the behavior of light at the first transition between media. Light scattering through heterogeneous media, as in Figure \ref{fig:refl}b is a higher-order phenomena.} The Fresnel equations relate the angles of reflection (e.g., the zenith component of surface normal) with the refractive index of the medium as well as polarimetric properties. 

\paragraph{Shape from Polarization (SfP)} is the term used in computer vision for a technique that estimates surface normals using the principles of the Fresnel equations. Classical SfP requires measurement of the polarimetric properties (through 3 polarized photos) and estimation of the refractive index to solve for the angle of reflection. We consider ourself with two primary branches of the SfP technique: first-order and higher-order. The first-order SfP techniques assume the reflection model akin in Figure \ref{fig:refl}a. Following this model has allowed shape estimation of
metals~\cite{morel2005polarization}, transparent objects~\cite{saito2001measurement,miyazaki2004transparent}, dark objects~\cite{miyazaki2012polarization}, and even ocean waves~\cite{zappa2008retrieval}. Higher-order SfP techniques rely on multiple interactions of light with a medium, as in the case of Figure~\ref{fig:refl}b. In such case, the Fresnel equations are applied differently, 
allowing for shape recovery of diffuse, subscattering surfaces. This is described with compelling experimental support by Atkinson and Hancock~\cite{atkinson2006recovery}.

\paragraph{Diffuse-specular separation} refers to a broad class of computational and optical techniques to decompose an image into a specular-only image and diffuse-only image. The most general techniques use only image or color information, but these can be susceptible to artifacts. For example, Nishino et al. introduced a technique that uses view-independent effects to identify the diffuse reflection in an image~\cite{nishino2001determining}. Other strategies combine image-based measurements with color analysis~\cite{nishino2001determining,lin2001separation,lin2002diffuse,tan2004separating,tan2005separating}. In this paper we show that it is beneficial to leverage the behavior of polarization to perform this separation. Previous attempts have used polarization to separate diffuse and specular components in the context of active illumination. In particular, spherical gradient illumination has been used by Ma et al.~\cite{ma2007rapid} and Ghosh et al.~\cite{ghosh2011multiview} to achieve photorealistic geometric reconstructions. For passive conditions, Nayar et al. introduced a separation technique that uses polarization images and color cues~\cite{Nayar97}. While successful, the Nayar method, and a related method proposed by Zickler et al~\cite{zickler2006reflectance} are limited by smoothness assumptions. In crux, though it is possible to directly combine existing work---for example, the SfP paper in~\cite{miyazaki2003polarization} uses the separation method from~\cite{tan2005separating}---we show that joint incorporation of reflection separation with SfP physics results is a beneficial strategy for addressing the mixed reflection problem in SfP.

\paragraph{Extended topics in polarization} that are tangentially relevant, but outside the direct scope of this paper are described in brief. While our paper considers linear polarization effects, 
work from~\cite{guarnera2012estimating} demonstrates shape reconstructions using circular polarization. Polarization information need not only be used for shape: prior art has considered problems like image dehazing~\cite{schechner2001instant}, illumination multiplexing~\cite{cula2007polarization}, panoramas~\cite{Schechner05}, underwater scattering~\cite{Treibitz09} or 3D displays~\cite{Lanman11}. In comparison to these related works, our paper is specific to the SfP problem. Future work could use, for example, the descattering model of~\cite{schechner2001instant}, to possibly obtain shape in scattering environments.  

\begin{figure}[t]%
\centering
\includegraphics[width=0.75\columnwidth]{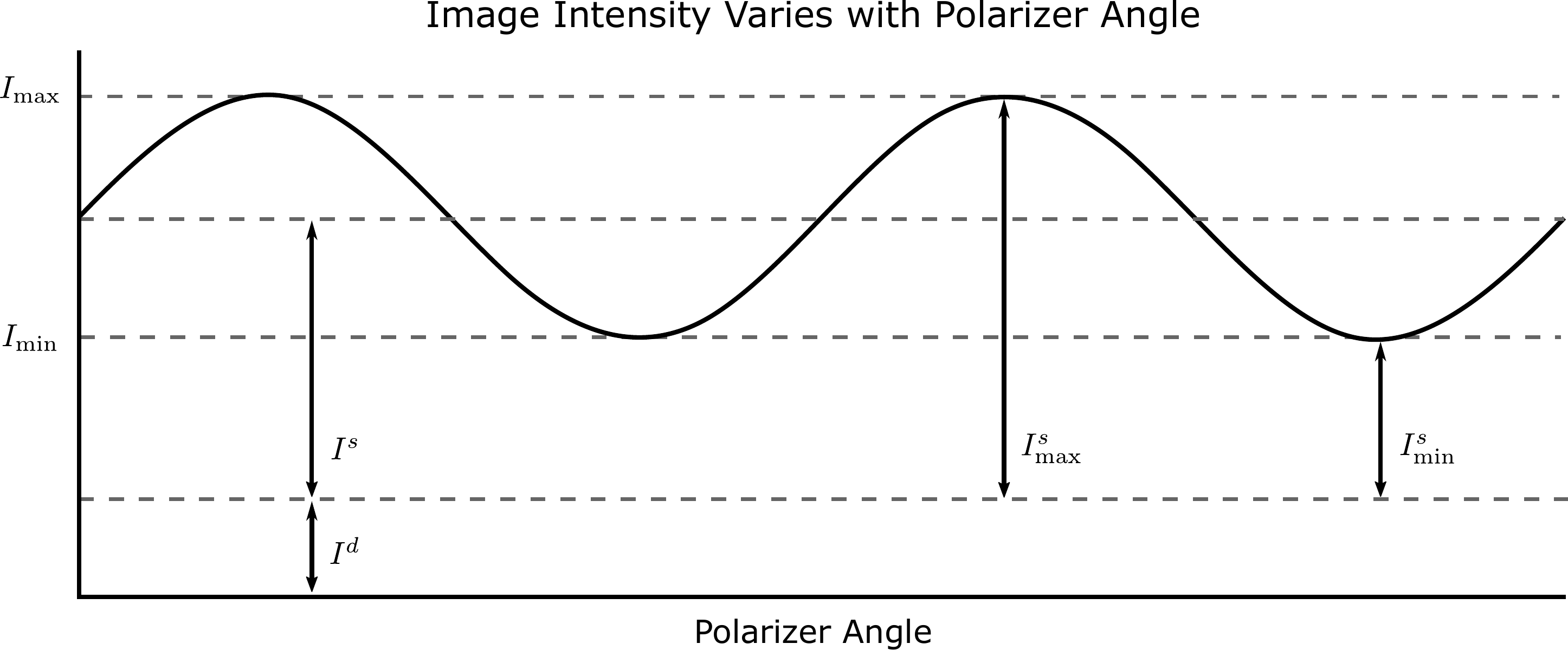}%
\caption{Notation used for intensity components. As the polarizer angle is rotated, the image intensity varies in accordance with Equation \ref{eq:formation}. Only the intensity quantities $I_{\max}$ and $I_{\min}$ can be directly measured with a camera. One of the aims of this work is to recover $I^{d}$ using the physical behavior of polarization, shape, and reflectance.}%
\label{fig:curves}%
\end{figure}


\section{Image Formation Model} 

\newcommand{\polangle}{\mathbf{\phi}_{\text{pol}}}

This section describes SfP in condensed form. Conceptually, SfP uses image-based measurements
to estimate the surface angles of azimuth, $\varphi$, and zenith, $\theta$. We will denote $\widehat \varphi$ and $\widehat \theta$ to represent
estimates of ground-truth quantities. 

The measured irradiance at a single scene point
is expressed as
\begin{equation}
I\left( \polangle \right) = \frac{{{I_{\max }} + {I_{\min }}}}{2} + \frac{{{I_{\max }} - {I_{\min }}}}{2}\cos \left( {2\left( \polangle - \phi \right)} \right), 
\label{eq:formation}
\end{equation}
where $\phi$ is the phase angle, and $I_{\max}$, $I_{\min}$ are the quantities shown in Fig. \ref{fig:curves}. Since a sinusoid has three unknowns, by sampling three different values of $\polangle$ it is possible to estimate $\phi$, $I_{\max}$, and $I_{\min}$.\footnote{If it is difficult to rotate the polarizer precisely, the recent method from~\cite{polcalibrate} allows determination of the angles from image-based measurements.} From these measurements, as detailed below, separate mechanisms are used to obtain the azimuth or elevation angles.

\subsection{Challenges with estimating azimuth angle} 

To obtain an estimate of the azimuthal angle, $\widehat \varphi$, early work in SfP has used a specular reflection model, as illustrated in Figure~\ref{fig:refl}a~\cite{wolff1991constraining,rahmann2001reconstruction,miyazaki2003polarization}. The maximum value of reflected light will occur when the light that reflects is perpendicularly polarized (since the Fresnel reflection coefficient for perpendicularly reflected light is greater). Then, since the maximum value of the cosine occurs at the origin, the azimuth angle is estimated as $\widehat \varphi = \phi$. 

Atkinson and Hancock later introduced a compelling technique to recover object shape from a diffuse reflection model, as illustrated in Figure~\ref{fig:refl}b. For such a scenario, it was observed that the direction of light propagation is reversed: light is refracted from the surface to air~\cite{atkinson2006recovery}. Since the direction of propagation is flipped, the minimum irradiance is now of interest, resulting in a shift in the estimated azimuth angle of $\pi /2$ radians, i.e., $\widehat \varphi = \phi \pm \pi /2$. 

Two key challenges occur with azimuthal estimation. First, since Equation \ref{eq:formation} includes a factor of 2 within the cosine, two azimuth angles, shifted apart by $\pi$ radians, cannot be distinguished in the polarized images.\footnote{Concretely, $\widehat \varphi$,and $\widehat \varphi + \pi$ return the same value for Equation \ref{eq:formation}} This first fundamental ambiguity is termed a \textbf{azimuthal ambiguity}, and applies to all SfP techniques. Second, for a general surface, not known \emph{a priori} to be diffuse or specular it is ambiguous as to whether the estimated angle should be shifted by $\pi /2$ radians or not. This second ambiguity, due to the surface reflectance, is termed as \textbf{azimuthal model mismatch}, and applies critically to our problem of mixed reflections.

\subsection{Challenges with estimating zenith angle}

Estimation of zenith angle relies on the degree of polarization, calculated as 
\begin{equation}
\rho = \frac{I_{\max} - I_{\min}}{I_{\max} + I_{\min}}.
\label{eq:dop}
\end{equation}
As in the case of azimuthal estimation, the type of reflection influences the reflection model. First, consider the specular model in Figure \ref{fig:refl}a. Substituting the Fresnel equations (see~\cite{hecht2002optics}) into Equation \ref{eq:dop} allows the degree of polarization to be written as
\begin{equation}
\rho = \frac{2\sin\theta\tan\theta\sqrt{n^{2}-\sin^{2}\theta}}{n^{2}-2\sin^{2}\theta+\tan^{2}\theta}
\label{eq:rhospec}
\end{equation} 
where $n$ denotes the refractive index and $\theta$ is the zenith angle. If some knowledge of $\rho$ and $n$ is obtained, then it is possible to solve Equation \ref{eq:rhospec} for an estimate of the zenith angle, $\widehat \theta$. This method is well-suited for highly specular objects and has been successfully used to estimate shape of metallic surfaces~\cite{morel2005polarization}. 

For diffuse reflections, as illustrated in Figure~\ref{fig:refl}b, the Fresnel equations are once again combined with the degree of polarization. However, this is performed for the model where light is transmitted from the surface to air, such that the relation is now expressed as
\begin{equation}
\rho  = \frac{{{{\left( {n - \frac{1}{n}} \right)}^2}{{\sin }^2}\theta }}{{2 + 2{n^2} - {{\left( {n + \frac{1}{n}} \right)}^2}{{\sin }^2}\theta  + 4\cos \theta \sqrt {{n^2} - {{\sin }^2}\theta } }}.
\label{eq:rhodiff}
\end{equation}

This work addresses two problems with zenith estimation. First, it is difficult to obtain an estimate of refractive index, that is accurate, at each pixel. If an improper refractive index is used, an error in zenith estimation occurs, termed in previous work as \textbf{refractive distortion}~\cite{kadambi2015polarized}. Second, it can be hard to know whether to use the model for a specular surface (Equation~\ref{eq:rhospec}) or a diffuse surface (Equation~\ref{eq:rhodiff}). Only in ideal scenarios do surfaces conform to specular and diffuse models -- real-world surfaces exhibit mixed reflections. This second source of error is referred to as \textbf{zenith model mismatch} in this paper. 


%

\section{Solving Model Mismatch}

To solve model mismatch error, consider the dichromatic reflection model, where the radiance from a single scene point is expressed as
\begin{equation}
I = I^d + I^s,
\end{equation} 
where $I^d$ and $I^s$ refer to the radiant intensity of diffuse and specularly reflected light. The prior work in SfP calculates $\rho$ from $I_{\text{max}}$ and $I_{\text{min}}$, the maximum and minimum intensities observed when rotating the polarizer. Following the work of Nayar et al.~\cite{Nayar97}, it can be assumed that only the specular component causes appreciable variation, such that the measured degree of polarization for a mixed surface is expressed as
\begin{equation}
\rho = \frac{I_{\max}^{s} - I_{\min}^{s}}{I_{\max}^{s} + I_{\min}^{s} + I^{d}} = {\widetilde \rho} \frac{I^{s}}{I} = {\widetilde \rho} \frac{I - I^{d}}{I},
\label{eq:rhotilde}
\end{equation}
where $I_{\max}^{s}$ and $I_{\min}^{s}$ denote maximum and minimum irradiance observed from specularly reflected light (see Figure \ref{fig:curves}). Under this simplification, it is possible to substitute Equation \ref{eq:rhospec} into Equation \ref{eq:rhotilde} to express the diffuse intensity as
\begin{equation} 
I^{d} = I(1- {\rho} \frac{n^{2}-2\sin^{2}\theta+\tan^{2}\theta}{2\sin\theta\tan\theta\sqrt{n^{2}-\sin^{2}\theta}}).
\label{eq:diff}
\end{equation}
This equation contains two unknowns: the intensity of diffuse reflection, $I^{d}$ and the refractive index $n$. Under a Lambertian approximation, the former quantity is constant across
different viewpoints. Additionally, since the refractive index is a physical property of the material, it is also constant across different viewpoints. The proposed strategy is to estimate the quantities on the right-hand-side of Equation~\ref{eq:diff} at different viewpoints, such that
\begin{equation}
{I^d} = f\left( {n,{\theta _i},{I_i},{ {\rho}_i}} \right) \triangleq I_i(1 - {\rho}_i \frac{{{n^2} - 2{{\sin }^2}\theta_i  + {{\tan }^2}\theta_i }}{{2\sin \theta_i \tan \theta_i \sqrt {{n^2} - {{\sin }^2}\theta_i } }}), 
\end{equation}
for the i-th view of $N$ total views. To recover $I^d$ a non-linear least squares problem can be solved of the form
\begin{equation}
\left\{ \widehat{I}^d, \widehat{n} \right\} = \argmin{I^d,n} \sum\limits_{i=1}^N (I_{d} - f(n, \theta_{i}, I_{i}, {\rho}_{i}))^2. 
\end{equation}
In this paper a sequential quadratic program is used to perform the minimization. Please refer to the supplement for implementation details.

\begin{figure}[t]%
\centering
\includegraphics[width=0.97\columnwidth]{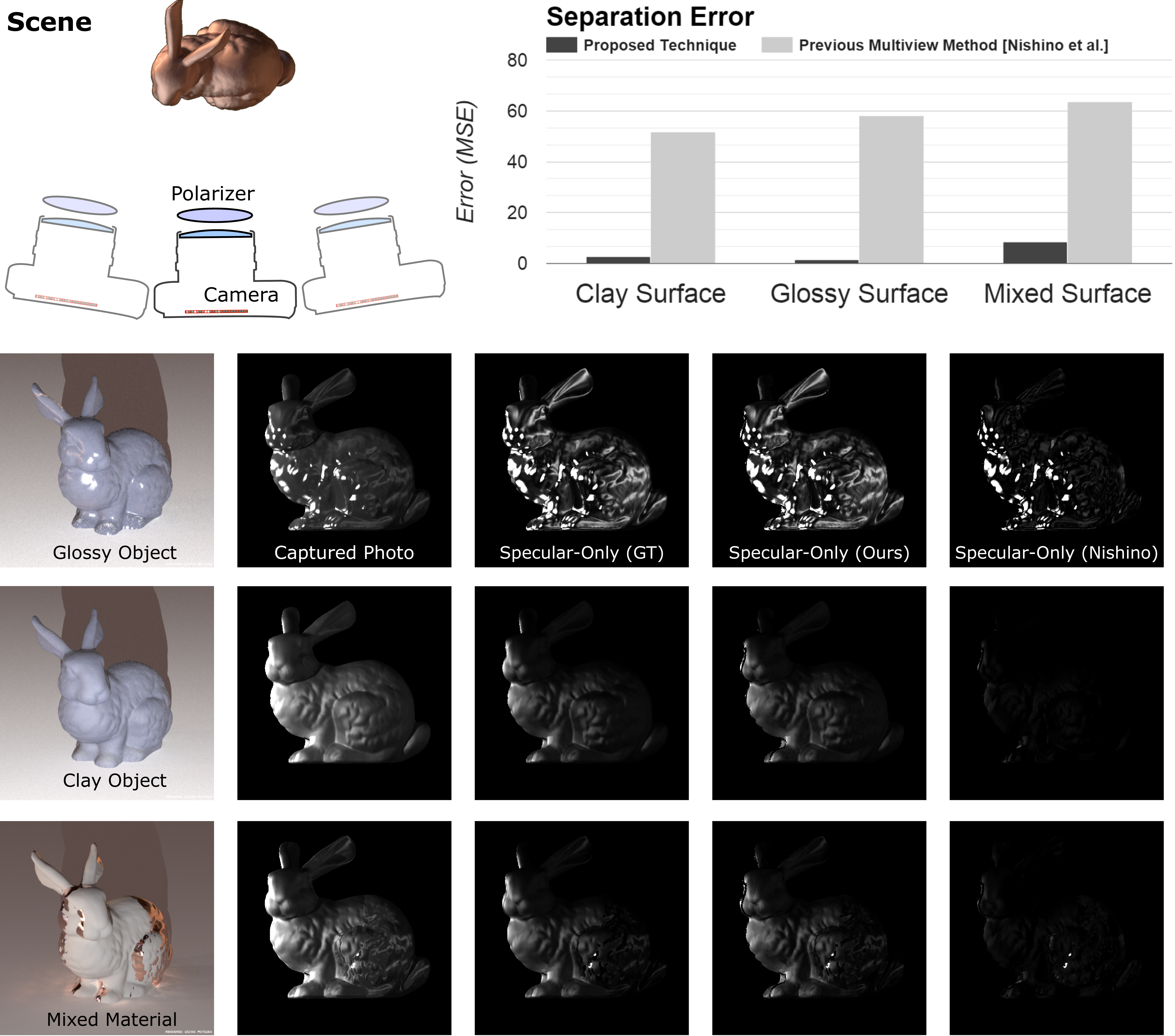}%
\caption{The proposed approach is able to separate reflectance for a variety of object textures (simulated example). Using the Mitsuba raytracer we render the Stanford bunny from three viewpoints, under three different material conditions (diffuse, glossy, and a spatially varying texture). The proposed technique is quantitatively compared with the previous work of~\cite{nishino2001determining}. By incorporating additional polarization, we demonstrate a significant reduction in error.}%
\label{fig:bunny}%
\vspace{-2mm}
\end{figure}


\section{Experimental Results}

Reflection models for SfP are not geared to handle mixed reflections. Existing solutions use a sequential approach: first, a robust algorithm is used to separate reflection components, following which SfP is performed. We provide a comparison to this sequential approach, using the multiview, reflection separation technique of Nishino et al.~\cite{nishino2001determining} as our point of comparison.

\paragraph{Implementation details:} All simulations were performed using the Mitsuba raytracer~\cite{jakob2010mitsuba}. The raytracer has been modified to acquire depth information and includes a Matlab script to simulate polarization measurements. The object remains static throughout all experiments as viewpoint diversity was acquired by moving the camera. Physical experiments
were performed with a Canon Rebel T3i camera with EF-S 18mm-55mm f/3.5-5.6 IS II SLR lens and a linear polarizer with quarter-wave plate, model Hoya CIR-PL. Three viewpoints were collected at 10 degree increments.

\paragraph{Diffuse-specular separation:} As shown in Figure \ref{fig:bunny}, the Stanford bunny is rendered with three different materials: clay, gloss, and a mixture of clay and gloss. Reflection separation is shown for the specular image component, for both the proposed technique and Nishino's method. Both techniques recover specular outliers, but the proposed technique recovers detail in regions that are of moderate specularity. As illustrated in the bar graph, the quantitative error is lower, for the proposed method, for all tested material configurations. Because the proposed
technique relies on viewpoint artifacts, classic artifacts like occlusions or lack of texture can lead to registration issues. This explains why our proposed method performs worse on the mixed material bunny (although the result is an improvement over Nishino's method).   

\begin{figure}[t]%
\centering
\includegraphics[width=1.0\columnwidth]{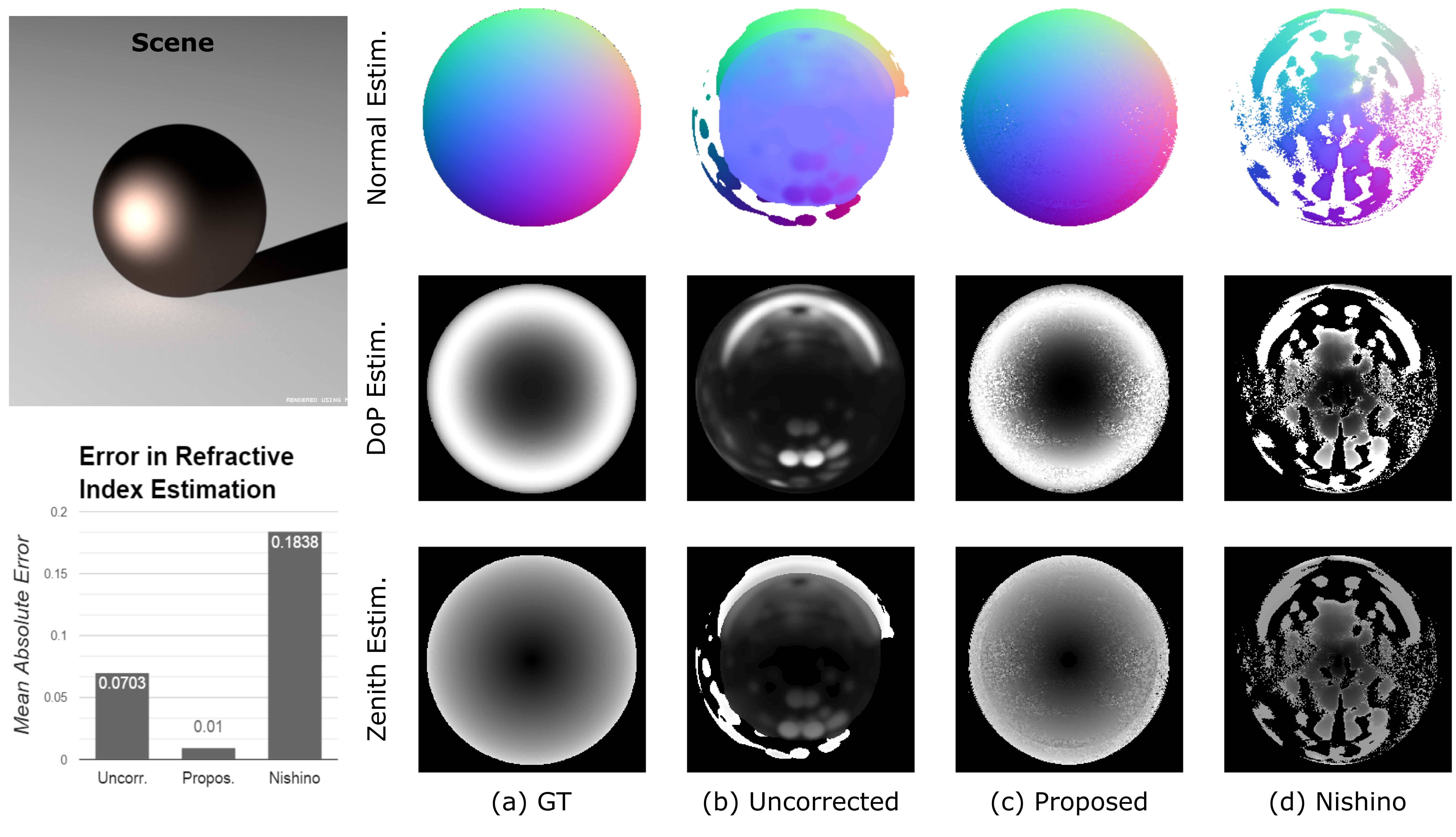}%
\caption{A glossy sphere is neither diffuse nor specular (simulated rendering). (a) The ground truth surface normal, degree of polarization and zenith angle. (b) Without correction, the estimates of surface normal, degree of polarization and refractive index (see bar graph) are incorrect. (c) The proposed method recovers a more accurate normal map, degree of polarization and refractive index than (d) the previous method of Nishino et al.~\cite{nishino2001determining}.}%
\label{fig:spheresim}%
\vspace{-2mm}
\end{figure}

\paragraph{Surface normal recovery:} Figure~\ref{fig:spheresim} uses a rendered sphere to show that the recovered surface normals are not accurate using naive shape from polarization (Figure~\ref{fig:spheresim}b. Simple pre-processing with Nishino's method, as shown in Figure~\ref{fig:spheresim}d, does not allow for robust surface normal recovery. The Nishino method, as a general method that does not account for polarization information recovers a the degree of polarization that does not conform with the physical scene. This leads to a poor estimate of zenith angle, and ultimately, surface normals. In comparison, the proposed technique, shown in Figure~\ref{fig:spheresim}c, shows clear recovery of the surface normals, as well as the degree-of-polarization anzenith angle. 

\paragraph{Refractive index recovery:} A benefit of the proposed technique is the ability to simultaneously recover per-pixel refractive index. The rendered sphere in Figure~\ref{fig:spheresim}
has a ground-truth refractive index of 1.5. The proposed technique estimates the refractive index as 1.49, for a mean absolute error of 0.01. Without applying the proposed correction for
mixed reflections, the error in refractive index estimation is 0.07. Interestingly, pre-processing with the Nishino method leads to a much greater error in estimating refractive index (0.18).

\begin{figure}[t]%
\centering
\includegraphics[width=0.99\columnwidth]{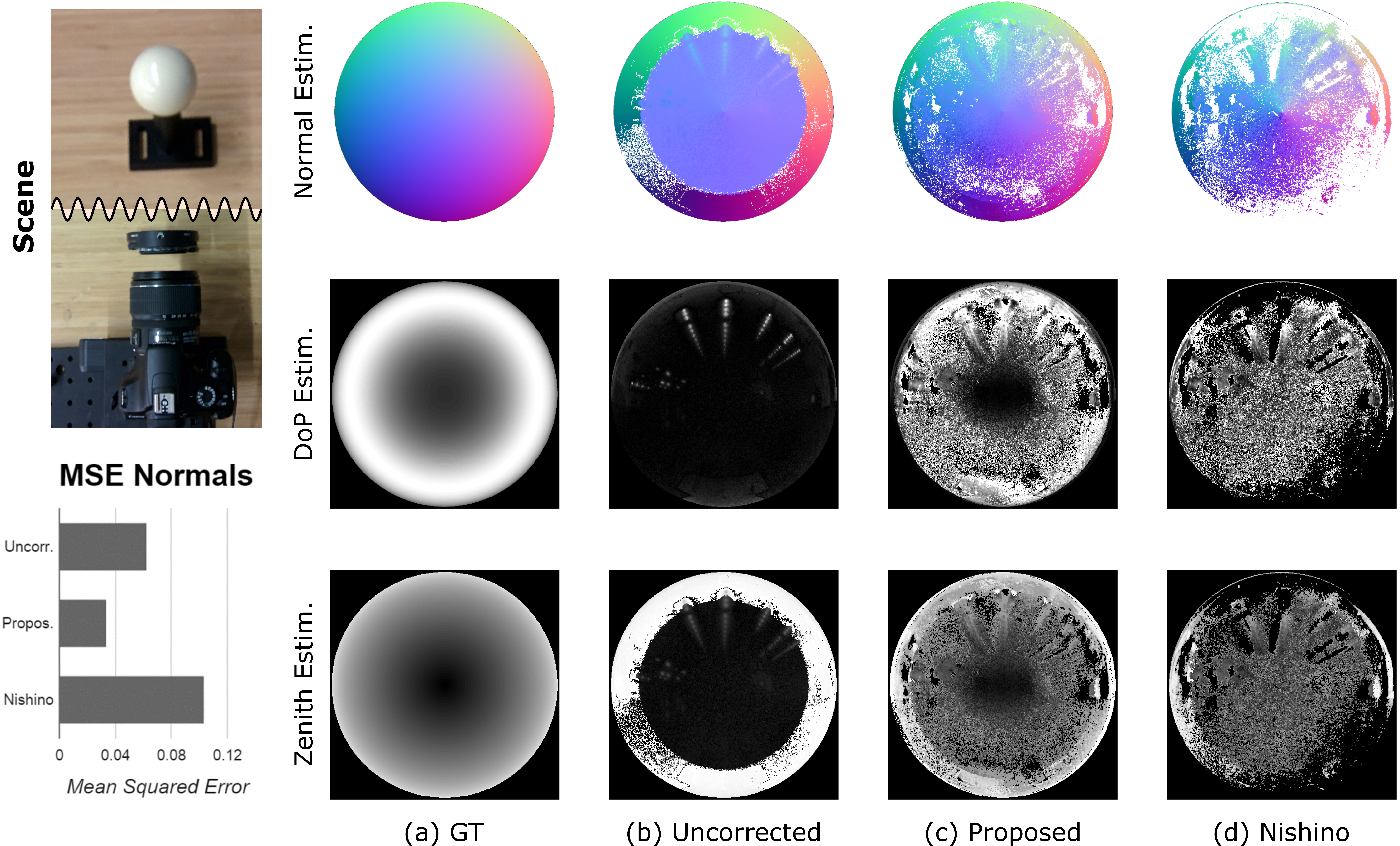}%
\caption{Validating the proposed technique with a physical experiment and comparisons to~\cite{nishino2001determining}. The uncorrect SfP result leaves something to be desired
as the normal map is noisy. The proposed correction algorithm reduces the MSE, while applying the technique from~\cite{nishino2001determining} increases the error.}%
\label{fig:spherereal}
\vspace{-2mm}%
\end{figure}

\paragraph{Physical experiment:} To validate our technique in the wild, a physical scene was set up in a similar fashion to the simulated examples. As illustrated in Figure~\ref{fig:spherereal},
a camera and polarizing filter are placed 50cm in front of a glossy sphere. Three viewpoints, at 10 degree increments, were captured. At each viewpoint, three polarized photos
were captured, for a total of nine photographs. The uncorrected surface normals, obtained from naive SfP, are poor. In particular, note the reflections of the ceiling lights in the normal map.
The proposed technique mitigates this issue, and reduces the MSE from 0.06 to 0.03. The Nishino method, while it does mitigate the dramatic specular reflections
from ceiling lights, results in a greater MSE. This is likely due to the significant holes in the normal map. 
 
%
%
%
%

\section{Discussion}

In summary, we have proposed a new technique to separate reflection components from a scene using a combination of passive polarization and viewpoint. To our knowledge,
this is the first paper to do so. While there are alternate ways to address the reflection separation problem -- for instance, through the use of color channels -- we believe that 
the proposed technique is a complementary approach that can be combined with previous methods. 

\paragraph{Benefits:} The proposed technique may find direct application in improving the quality of SfP and related algorithms. Prior art in SfP has not analyzed in-depth the 
impact of mixed reflections. This paper has shown that it may not be sufficient to sequentially apply an existing algorithm to first separate reflection components. Rather, the 
physics of polarization that are used to obtain shape, can \emph{also} be used to separate reflection components. With the increased interest in multiview methods (e.g. KinectFusion~\cite{Izadi2011KinectFusion}), it seems logical to consider the inclusion of the proposed technique within such frameworks. In addition, the proposed technique has shown recovery of refractive index, which is a challenging problem in computational imaging often addressed with calibrated optical setups~\cite{wetzstein2011hand}. The ability to estimate refractive index is shown to greatly improve the accuracy of SfP, but may also find use in other applications like object detection.

\paragraph{Limitations:} We follow previous work in using the unpolarized world assumption -- the light incident on an object is initially unpolarized. In scenes with significant specular reflections -- like a house of mirrors -- the unpolarized world assumption is violated. It should be noted that prior art has empirically observed the validity of the unpolarized world assumption in realistic
scene conditions~\cite{kadambi2015polarized}. Although our paper also acquires refractive index, for the sole purpose of reflection separation, other strategies that use fewer images (e.g. Tan et al.~\cite{tan2005separating}) may be preferable. Specifically, the proposed technique uses three polarized images at a minimum of two viewpoints -- a minimum total of 6 images are required. However,
the intended application of this technique is to SfP, where it is expected to capture multiple images, and where it is desirable to estimate the refractive index. 

\paragraph{Open challenges:} While the proposed technique forges a strong link betweeen shape, passive polarization, and reflectance, several open topics remain. For example, 
would the technique improve if more viewpoints were captured; would circular polarization allow for more information to be gleaned; and could this method be combined with other frameworks (for example, color, viewpoint, and polarization)? In conclusion, we hope that this paper may improve the practicality of SfP, allowing surface normals to be estimated on surfaces
with mixed reflective properties. 

\clearpage

\bibliographystyle{splncs}
\bibliography{egbib}
\end{document}